# Can Large Language Models Code Like a Linguist?: A Case Study in Low Resource Sound Law Induction

Atharva Naik[1]  Kexun Zhang[1]  Nathaniel Robinson[2]  Aravind Mysore[1]  Clayton Marr[3]
Hong Sng[1]  Rebecca Byrnes[1]  Anna Cai[1]  Kalvin Chang[1]  David Mortensen[1]
Carnegie Mellon University[1]  John Hopkins University[2]  Ohio State University[3]
{arnaik,kexunz,amysore2,hongsng,rbyrnes,annacai,dmortens}@cs.cmu.edu
nrobin38@jhu.edu  kalvinc@alumni.cmu.edu

## Abstract

Historical linguists have long written a kind of incompletely formalized "program" that converts reconstructed words in an ancestor language into words in one of its attested descendants that consist of a series of ordered string rewrite functions (called SOUND LAWS). They do this by observing pairs of words in the reconstructed language (protoforms) and the descendent language (reflexes) and constructing a program that transforms protoforms into reflexes. However, writing these programs is error-prone and time-consuming. Prior work has successfully scaffolded this process computationally, but fewer researchers have tackled SOUND LAW INDUCTION (SLI), which we approach in this paper by casting it as Programming by Examples. We propose a language-agnostic solution that utilizes the programming ability of Large Language Models (LLMs) by generating Python sound law programs from sound change examples. We evaluate the effectiveness of our approach for various LLMs, propose effective methods to generate additional language-agnostic synthetic data to fine-tune LLMs for SLI and compare our method with existing automated SLI methods showing that while LLMs lag behind them they can complement some of their weaknesses.

## 1 Introduction

In the 19th century, European linguists, then called "philologists," made a striking discovery: the sounds of the words of languages change in a principled, rule-governed way. When formulated, these rules are called SOUND LAWS. Linguists determined that the ancestral forms (protoforms) of words could be reconstructed by comparing their descendant words (reflexes) across a language family, a set of languages sharing a common ancestor. The words in any member of the language family could then be derived by applying a cascade, or ordered sequence of sound laws (Marr and Mortensen, 2020), to the protoforms in the reconstructed ancestor language. In writing these cascades of sound laws, Neogrammarian linguists were essentially writing programs—composed sequences of regular expression replacements like

$$\emptyset > \text{k}/\_\_\left\{\begin{array}{c} \text{i} \\ \text{u} \end{array}\right\}\#$$

("rewrite the empty string with /k/ after /i/ or /u/ at the end of a word", sound change in the history of Huishu).

Computational linguists have made great progress in solving the first problem (reconstructing ancestral words based on modern words, or BACKWARDS RECONSTRUCTION) with probabilistic graphical models (Bouchard-Côté et al., 2013; Ciobanu and Dinu, 2014; Hruschka et al., 2015) and neural methods (Meloni et al., 2021; Kim et al., 2023; Lu et al., 2024a; Hartmann, 2019, 2021; Lu et al., 2024a,b; Fourrier, 2022). This is significant, since this task requires enormous cognitive and temporal resources and is incredibly error-prone. The second problem, inducing sound changes from sets of ancestor-descendant pairs or, (COMPUTERIZED) FORWARD RECONSTRUCTION (Sims-Williams, 2018) has proved more elusive (Luo, 2021) and the most successful efforts so far have been rule-based (Marr and Mortensen, 2020, 2023; Bodt and List, 2022).

If the COMPUTATIONAL FORWARD RECONSTRUCTION (CFR) problem is essentially the problem of writing a (conceptually simple) program (Sims-Williams, 2018), all approaches to this problem are, at some level, program synthesis. Inducing a single sound law is discovering a string-rewrite function; inducing a cascade of sound laws is discovering a composition of such functions such that input ancestral words are deterministically converted into descendant words. Put differently, the forward reconstruction problem is a problem of Programming by Example.

There are existing robust approaches to program synthesis and programming by example. Since Large Language Models (LLMs) have been successfully applied to this kind of program synthesis before (Austin et al., 2021; Lahiri et al., 2022; Jain et al., 2022), we propose PySLICoder, a method for automated synthesis of Python code representing sound laws using LLMs to support CFR.

Our contributions are as follows:

1. A code-generation-from-examples formulation of sound law induction (SLI) to leverage advancements in program synthesis.
2. An LLM based approach for SLI as Python code generation without any language-specific data.
3. Benchmarking GPT-4 and various open source LLMs for SLI.
4. Synthetic data generation methods to improve the performance of open source LLMs for SLI.
5. Comparison of LLMs with other learning-based SLI approaches.

## 2 Related Work

### 2.1 Automatic sound law induction

List (2019) automatically induced sound correspondences with a minimum clique cover algorithm, but these correspondences lacked the contexts associated with actual sound laws. Chang et al. (2023) apply Albright and Hayes (2003)'s deterministic string algorithm (Wilson and Li, 2021) to automate sound law induction. While Marr and Mortensen (2020) propose the task of computerized forward reconstruction (CFR) and Luo (2021) proposes a Monte Carlo Tree Search (MCTS) algorithm for automatic derivation of sound laws, our approach performs both CFR and sound law induction with greater sample efficiency. More broadly, to our knowledge, we are the first to perform forward reconstruction using large language models.

### 2.2 Linguistic Reasoning with Large Language Models

LLMs have a mixed record when it comes to linguistic reasoning. ChatGPT generalized morphological patterns to nonce words more poorly compared to humans (Weissweiler et al., 2023). GPT-3.5-turbo-instruct could analyze compositional word formation and derivation in German compounds but could not identify illicit derivations (Weller-Di Marco and Fraser, 2024). Zhou et al. (2024) found that LLMs are not good at classifying so-that constructions. However, LLMs demonstrate lexical-syntactic flexibility in that (Mortensen et al., 2024). Whereas prior work focused on morphology, syntax, and semantics, we are the first to use LLMs for historical linguistics, generally, and historical phonology, specifically.

### 2.3 Programming by Example

Code generation from input-output examples that demonstrate the expected behavior is well explored in the literature of programming by examples (PBE) (Gulwani, 2016). Researchers in this space have explored ways of leveraging input-output examples to guide programming synthesis in neural models (Ye et al., 2021; Shi et al., 2023) with recent efforts focused on prompting LLMs with examples or test-case information for program synthesis (Austin et al., 2021; Lahiri et al., 2022; Jain et al., 2022). Austin et al. (2021) benchmark LLMs for program synthesis in Python from input-output examples and natural language (NL) intent, by proposing the Mostly Basic Programming Problems (MBPP) dataset. Lahiri et al. (2022) propose an interactive approach for formalizing underspecified intents by obtaining user feedback on proposed input-output examples. Contemporary work has shown that LLMs are not effective at PBE by default but can be improved by fine-tuning as long as the train and test distribution overlap (Li and Ellis, 2024).

Within the realm of linguistic reasoning (Vaduguru et al., 2021) have shown that PBE techniques can be leveraged for sample efficient linguistic generalizations using FlashMeta (Polozov and Gulwani, 2015) a data-driven PBE framework. Moreover Luo's MCTS algorithm for sound law induction (Luo, 2021) essentially generates simple regular expression programs to represent sound laws. Motivated by these approaches and the observation that LLMs when coupled with fine-tuning can support PBE, we propose synthetic data generation and fine-tuning methods for sound law induction.

## 3 Method

We propose a prompting-based approach with examples and instructions shown to the LLM to sample multiple Python sound law programs which are then executed on the examples as test cases to identify functionally correct sound laws. For the discovery of the cascade of chronologically ordered sound laws, we employ a beam search approach with an edit distance-based reward (described in section 4.3). Additionally, we also develop multiple algorithms for generating various kinds of language-agnostic synthetic data to teach the language model basic string manipulation and more linguistically motivated sound law production. We perform supervised fine-tuning (SFT) on Magicoder (Wei et al., 2024) an open-source language model with the synthetic data to create PySLICoder. Finally, we propose techniques like inference time example selection (ITES) and ensembling of code generators to further boost the sound law induction performance of PySLICoder. Each of these methods is discussed in detail in the subsequent sections.

### 3.1 Sound Law Induction Prompting

We prompt all the LLMs including PySLICoder with the tokenized (using PanPhon (Mortensen et al., 2016)) protoforms and reflexes as the input and output respectively to generate sound laws as shown in Figure 1. The prompt contains six main components including instructions (red), code (yellow), and examples (green). The first and second sections describe the task, evaluation measures, and the BasicAction class we use to represent sound laws as Python code. The third section represents the protoforms and reflexes and their tokenized versions. The fourth and fifth section contain code examples demonstrating usage of the BasicAction and the source definition for it. Finally, the sixth and last section contains additional instructions like avoiding importing other packages and not modifying or repeating the definition of BasicAction.

The BasicAction class and its input specification are described in greater detail in Appendix A.1. We sample $s$ (=20) sound laws and sort them based on the edit distance reward on the input and output examples. A perfect sound law (pass rate = 1) also achieves a perfect reward of 1 (by achieving the target output for each input) so we pick the sound law with the highest reward.

### 3.2 Sound Law Search

While the procedure described in the previous section is sufficient for uncovering single sound laws, the SLI task requires us to discover the cascade of all sound laws (or changes) in the order in which they occur. To achieve this we apply a beam search style algorithm as shown in Figure 3.

Let us say we have $k$ simultaneous hypotheses or beams, each representing a sound law cascade, and at a given step, each beam is *expanded* by sampling $s$ new sound laws. We also limit the max iterations of the beam search (or the maximum number of sound laws in the cascade) to $m$.

At each *expansion step* we sample $s$ sound laws per hypothesis by prompting the LLMs for single sound law generation but with the examples having the intermediate input (output from last sound law in the cascade) and the target output, giving us $k \times s$ beams. After each expansion we *contract* the hypothesis space back to $k$ beams by picking the top-$k$ hypotheses based on the aggregated edit-distance reward (described in section 4.3). At the $m^{\text{th}}$ *contraction step* we pick the best beam out of the $k$ based on the reward.

### 3.3 Synthetic Data Generation

We propose two synthetic data generation algorithms to generate language-agnostic data of different complexities to provide additional task-specific supervision to the LLMs for the task. We outline these here.

**String manipulation sound laws (SMP)** This data simulates simple, randomly sampled sound laws with environments of one to three phones with a few of them (25%) having boundary based conditions (one of word start, word end, not word start and not word end), that perform one or more phone additions, deletions or substitutions. For each sound law we randomly sample $N$ (=50) protoforms such that at least $\frac{2}{5}N$ protoforms contain one or more occurrences of the environment with $\frac{1}{10}N$ each beginning and ending with the environment respectively, and $\frac{1}{10}N$ each containing at least one and at least two occurrences of the environment in the middle of the word respectively. The remaining $\frac{3}{5}N$ protoforms are sampled randomly. We obtain the reflexes by generating BasicAction code for the sound laws and applying them on the protoforms. We outline these steps in Algorithm 1.

**Linguistic sound laws (LING)** We also created synthetic data via an algorithm that simulates sound change at the granularity of articulatory features, like genuine sound changes. For example, in the history of English, a change occurred causing word-initial /k/ to be deleted before /n/ (so that /knixt/ 'knight' ultimately came to be pronounced

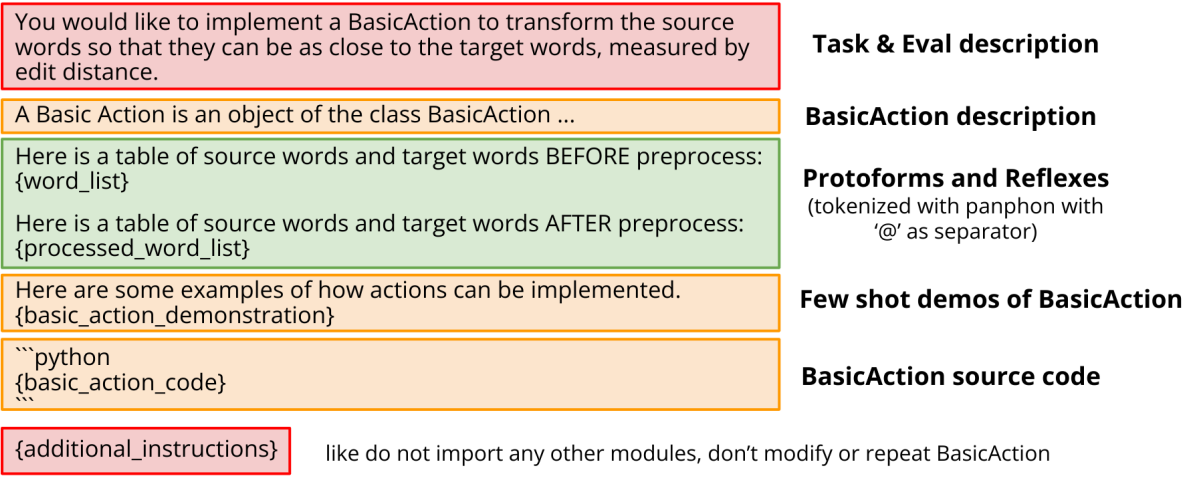

Figure 1: Prompt used for synthesizing single sound laws at each step from the protoforms or attested forms at the current step and the eventual targets or reflex forms.

/najt/). The algorithm we use accepts $N$ (=50) nonce words based on natural language as parent forms and then produces $k$ (=3) sound laws for each of $n$ (=2000) synthetic daughter languages. To create each of these $n$ (=2000) synthetic daughter languages, we generate nonce words in the International Phonetic Alphabet via Wuggy (Keuleers and Brysbaert, 2010) and Epitran (Mortensen et al., 2018), based on 50 randomly sampled words from one of seven languages (randomly selected). Then, to create each sound change rule, we (1) sample the length of the sequence on which the sound change operates, as well as the lengths of its left and right context requirements; (2) sample a vector of articulatory feature requirements for each phone in both the changing sequence and its contexts, which must be met for the change to occur; (3) sample the sound change itself by allowing a $\frac{1}{8}$ chance of a substitution or deletion of any given phone and a $\frac{1}{16}$ chance of insertion between any two phones. We only retain rules that actually apply to at least $m$ (=3) of the parent words (i.e. the nonce words). The details of this procedure are in Algorithm 2.

### 3.4 Supervised Fine-tuning (SFT) on Synthetic Data

For our proposed approach PySLICoder we use the open-source LLM Magicoder (Wei et al., 2024) as the base model for reproducibility. We picked Magicoder because it has decent zero-shot performance (pass-rate of 0.1154 and reward@1 of 0.2175) while having the least number of parameters (6.7B) among the LLMs benchmarked, making it less computationally expensive to fine-tune on task-specific data. We perform supervised fine-tuning (SFT) with a text completion objective (cross-entropy loss) on the synthetic data generated from SMP or LING. We use the fine-tuning code published by the Magicoder creators[1] and train the model using an 80GB, A100 GPU for three to five epochs, evaluating the model every 500 steps.

### 3.5 Inference Time Example Selection (ITES)

Usually, sound laws don't affect all the protoforms and reflexes given as examples to the LLM (because their structural conditions are not met by all examples). We leverage this fact to select the most important set of examples to be shown to the LLM for sound law induction, to reduce the size of the input and focus on the most informative cases. We achieve this by first traversing through the list of all examples and analyzing the ones where the input is not equal to the output, using edit distance analysis to figure the input phones likely to trigger the change and accumulating the minimal set of change-triggering characters for the sound law ($\Delta_{SL}$). Then we remove any examples where the input is the same as the output and the input doesn't contain any character from $\Delta_{SL}$. We observe that this method can boost the pass rate or the reward@k.

---

[1] https://github.com/ise-uiuc/magicoder/blob/main/src/magicoder/train.py

### 3.6 Ensembles of Code Generators

We can combine the results of two or more code-generating language models (say $\{LM_1, \ldots LM_k\}$) by pooling the candidate sound law programs ($P_{SL}^{(i)}$) produced by them for a given set of input-output examples $\bigcup_{1 \leq i \leq k} \{P_{SL}^{(1)}, \ldots P_{SL}^{(k)}\}$ and then picking the best sound law based on the edit-distance reward or pass rate. At the cost of greater inference time, the resulting ensemble of models is at least as good as the best-performing model and can be better than any of the models if they struggle with different kinds of sound law programs. We show in Table 1 that the PySLICoder ensemble of models trained with different kinds of synthetic data combined with Codestral (AI, 2024) can achieve performance close to GPT-4 for single sound law generation

## 4 Experiments

We perform experiments using three different kinds of datasets to determine the most effective code LLMs for SLI, evaluate their performance at uncovering simple multi-law cascades, and compare their effectiveness against traditional automated SLI methods for realistic low-resource sound change data. We prompt all the models identically with the prompt shown in Figure 1 which includes the source code of the BasicAction with demonstrations. The subsequent sections describe the datasets, baselines, and evaluation metrics.

### 4.1 Datasets

We create three types of evaluation datasets to evaluate the ability of LLMs to recover cascades of sound laws in different settings.

**Single Law Prediction:**
In this task, the LLM was tasked with inducing a single sound law to explain the development of input forms into their paired outputs. For these pairs, we recruit headword/reflex pairs from Huishu and its Tangkhulic relatives Ukhrul and Kachai for sets of paired Proto-Tangkhulic ancestral forms and reflexes representing a single sound change (**Mortensen, 2023**). This dataset corresponds to the 26 sound laws shown in Table 4, with 6-74 example lexemes per sound law (36 on average). We utilize this data as a launching bed to compare and benchmark the performance of various LLMs of code for SLI.

**Multi Law Synthetic Reflex Prediction:**
For this task, the LLM had to induce an ordered sequence of rules that, applied to the protoforms in order, would generate predicted outputs that to maximally match those synthetically generated by applying the subsampled rules from a traditionally constructed cascade in order. To generate data for this task, the traditional method was applied to manually produce a cascade for the development of Proto-Tangkhulic (Mortensen, 2023) into Huishu; this cascade was then recalibrated, simplified and improved in accuracy using the CFR and debugging system DiaSim (Marr and Mortensen, 2020). For the resulting cascade, written in the SPE format typical of phonology literature, a corresponding sequence of BasicAction-style rules were built. Evaluation data for this task consisted of sets of pairs of proto-forms with reflexes synthetically generated by subsampling a set of rules from the human-produced cascades described above, and applying them to each protoform in order to generate synthetic reflexes. Specifically, we generated 10 sets of 5 sampled rules to generate the synthetic reflexes to be paired with protoforms, choosing a set of 50 protoforms for each sampled rule cascade in a way that at least 50% of them remain unchanged after applying the rules, and evaluating the SLI performance of the best-performing LLMs in the single sound law setting. Thus, in essence, the LLM's task was to generate a cascade that produced the same results as these rules applied in order to the protoforms would, whether or not they do so by means of an identical cascade. While the reflexes were synthetically generated, this represents a more realistic scenario in that the history of the development of the "language" consists of more than a single sound change, and observing how those LLMs that performed best in the single sound law setting fared in this task.

**Final Reflex Prediction:**
In this task, the LLM's goal was to build a cascade of sound laws that, applied in order to each form in a proto-language, would generate predictions that most often matched their real observed reflexes in a descendant language. This task represents a typical SLI setting where we expect the LLMs to reconstruct the entire cascade of sound laws for a language's development. Data here consisted of headword/reflex pairs between Proto-Tangkhulic and Huishu (221 examples) drawn from Mortensen (2023), and between Proto-Polynesian, and respectively, its descendants Hawaiian (106 examples), Niue (130), Samoan (141), and Tongan (126), from Greenhill and Clark (2011). Since the Polynesian data contained numerous cases where the 'descendant' language word was not the *strictu sensu* reflex of the protoform, some adjustments were applied: descendant language words not actually descended in any way from the protoform (due to loaning, semantic shift, etc.) were simply removed, while protoform/descendant pairs that had seen affixes removed or added were fixed so that the pair had the same morphological structure, with cascades constructed and then simulated via DiaSim (Marr and Mortensen, 2020) consulted to gauge the plausibility of reflexes in light of the other items in the dataset for each language; in cases where forcing morphological equivalency between the protoform and reflex would have phonological side effects, the pair was pruned out. The number of examples listed in parentheses for each language represents the set of pairs after these adjustments had been implemented. Polynesian and Tangkhulic data are low-resourced compared to the Romance (e.g. Latin to Italian) used by prior automated sound law induction work (Luo, 2021), so by running the system of the aforementioned paper, we can compare our trained LLM against the performance of earlier automated SLI languages for this lower resource setting.

### 4.2 Baselines

We compare our approach with three kinds of modeling methods:

**Luo's MCTS:** Is a reinforcement learning-based approach with hierarchical MCTS planning where the states represent vocabularies and actions represent sound laws and the goal of the "agent" is to apply the sequence of sound laws to go from the initial state (ancestral form) to a final state close to the descendant forms while maximizing an edit distance based reward (defined in section 4.3)

**Open Source LLMs:** We evaluate various instruction-tuned LLMs of code in the scale of 6.7B-33B parameters as outlined below:

**CodeLlama-7B-Instruct** (Roziere et al., 2023): Is fine-tuned from the general purpose Llama-2 model (Touvron et al., 2023) with code data and an additional infilling objective.

**CodeQwen1.5-7B-Chat** (Bai et al., 2023): Is fine-tuned from the general purpose Qwen model.

**DeepSeekCoder-7B-Instruct-v1.5** (Guo et al., 2024): Is an open source LLM trained on high-quality project-level code with large context windows (16K tokens) and an infilling objective.

**Magicoder-6.7B-S-DS** (Wei et al., 2024): Is instruction tuned using synthetic data with Evol-Instruct and OSS-Instruct with CodeLlama (CL) or DeepSeekCoder (DS) as the base model. We chose the DS version because of better, reported performance on coding benchmarks.

**StarCoder2-15B-Instruct** (Lozhkov et al., 2024): Is a popular open-source LLM released by the Big-Code project, trained on the high-quality StackV2 dataset that includes GitHub pull requests, Kaggle notebooks, and code documentation.

**Codestral-22B-v0.1** (AI, 2024): Is an open-source code LLM recently released by Mistral AI that supports 80+ programming languages.

We also evaluate some general purpose open source LLMs with programming capabilities:

**Llama-3-8B-Instruct** (Team, 2024): Is the latest general purpose instruction tuned LLM in the Llama series which outperforms CodeLlama.

**Phi-3-14B-128k-Instruct** (Abdin et al., 2024): Is a general purpose LLM released by Microsoft that has performance comparable to and in some cases better than Mixtral-8x7B mixture of experts model (Jiang et al., 2024) and GPT-3.5 (Ouyang et al.), especially for coding benchmarks like MBPP (Austin et al., 2021) and HumanEval (Chen et al., 2021).

**Closed Source LLM APIs:** We utilize closed source LLMs like OpenAI's GPT-4 (Achiam et al., 2023) to potentially get an upper bound on the performance of LLMs for sound law induction and make a strong case for a code-generation-from-examples approach for it with a strong LLM.

### 4.3 Evaluation Metrics

We evaluate the performance of all approaches by executing the BasicAction programs generated by the LLMs to obtain "predicted reflexes" and compare them with the known reflexes using the metrics described below:

**Edit Distance Reward:** This metric is based on the reward function defined by (Luo, 2021). The original function computes the reward for multi-law sound law search as shown in the equation below, where the initial ancestral form of the word is denoted by $s^{start}$, the current form within an intermediate daughter language is $s$, the immediate descendant of $s$ is $s^{next}$ (obtained by applying the "ac-

tion" or sound change 'a'), and the eventual form is denoted by $s^{end}$:

$$R(s, a) = \frac{dist(s, s^{end}) - dist(s^{next}, s^{end})}{dist(s^{start}, s^{end})}$$

The distance function (*dist*) is the aggregated Levenshtein edit distance between all protoforms $s$ and reflexes $s'$ in a parent-daughter language pair.

$$dist(s, s') = \sum_{i=1}^{n_v} dist(s_i, s'_i)$$

We use a simplified version of the reward that compares the predicted reflexes, or eventually predicted reflexes in the multi-law setting, against the ground truth reflexes by setting $s = s^{start} = s^{source}$ and $s^{next} = s^{pred}$ for predicted reflexes and $s^{end} = s^{target}$ for the target reflexes.

$$R(s^{source}, s^{pred}, s^{target}) = 1 - \frac{dist(s^{pred}, s^{target})}{dist(s^{source}, s^{target})}$$

From the equation above it is clear that the metric attains a maximum value of 1 when $s^{pred} = s^{target}$ or $dist(s^{pred}, s^{target}) = 0$. Also, the reward can take negative values, when $dist(s^{pred}, s^{target}) > dist(s^{soruce}, s^{target})$. Additionally, we define a reward@m which evaluates the average reward achieved by the top-m hypothesized cascades (denoted by $\mathcal{R}^m$) (when sorted by the reward) and then aggregates it over all instances $i$ (sound laws, languages or language subsets) in the data ($\mathcal{D}$).

$$\text{reward@m} = \frac{1}{|\mathcal{D}|} \sum_{i \in \mathcal{D}} \sum_{s \in \mathcal{R}^m} R(s^{source}, s^{pred}, s^{target})$$

**Pass Rate** captures whether a model can produce a perfect rule, i.e. a rule that passes all test cases/explains all examples. It is the percentage of instances $i$ with reward@1 = 1 or:

$$\text{pass\_rate} = \frac{1}{|\mathcal{D}|} \sum_{i \in \mathcal{D}} \sum_{s \in \mathcal{R}^1} \mathcal{I}[R(s^{source}, s^{pred}, s^{target})]$$

Where $\mathcal{I}[x] = \begin{cases} 1, & \text{if } x = 1 \\ 0, & \text{otherwise} \end{cases}$

## 5 Results

### 5.1 Single Law Prediction

We show the Reward@m and pass rate obtained by various LLMs of code on single law prediction in Table 1. The pass rate is the most important metric here since it only rewards completely correct rules, i.e. all protoforms are transformed into the target reflexes. A high pass rate is required, since a rule even if slightly incorrect can propagate error in rule cascades during the prediction of multiple sound laws.

The results show that GPT-4 can achieve a perfect pass rate zero-shot while the smaller open source LLMs (6.7B-15B parameter range) fail to get a pass rate greater than 15% . The Codestral-22B model can get a much better pass rate of 54% but still lags behind GPT-4, while our method PySLICoder when fine-tuned on the SMP synthetic data can surpass Codestral with a 61% pass rate and when combined with ITES and ensembled with the LING SFT checkpoint can achieve a 65% pass rate. Finally, PySLICoder ensembled with Codestral can get a pass rate of 80%, which is much closer to GPT-4 with only 35.4B parameters.

### 5.2 Multiple Law Synthetic Reflex Prediction

We show the performance of various LLMs on multi-sound laws subsets of Huishu in Table 2. We use the sound law search algorithm mentioned in section 3.2 for all models with 20 beams (k=20) and one sample at each step (s=1) instead of 20 samples in the single law case. The metrics are computed for the final output generated by the rule cascade in each beam and the Reward@m denotes the average reward across m best beams. We observe that GPT-4 performs the best, however, among the open-source models, DeepSeek-Coder and Magicoder performs really poorly zero-shot, despite using the DeepSeekCoder 33B parameter variant. PySLICoder outperforms all zero-shot models except Codestral. Finally, we also compute the result of ensembling all LLMs and the results indicate a lot of scope for improvement and that there are some cases that none of the LLMs can address.

### 5.3 Final Reflex Prediction

We show the results in Table 3. The results show that Luo's MCTS model outperforms the LLMs on Reward@1 by a sizable margin but subsequent beam hypotheses produced by it degrade rapidly in quality as shown by the other Reward@m scores. Additionally, we notice very little variation in performance between the LLMs with Codestral being the best followed by PySLICoder and GPT-4 surprisingly having the worst performance (please note that we ran GPT-4 on a subset of the data due to API cost constraints). When looking at the performance breakdown per language shown in Table 5 we notice that for the LLMs, Tongan and Huishu are the most challenging languages while Hawaiian is the easiest. However, for Luo's method, Niue turns out to be a major point of failure while it can do well on Huishu, indicating complementary points of failures for both kinds of approaches.

| Model | Reward@1 | Reward@3 | Reward@5 | Reward@10 | Pass Rate |
|---|---|---|---|---|---|
| CodeLlama-7B-Instruct† | 0.0769 | 0.0256 | 0.0154 | 0.0070 | 0.0769 |
| CodeQwen1.5-7B-Chat† | 0.21 | 0.0956 | 0.0574 | 0.0231 | 0.1154 |
| DeepSeekCoder-7B-Instruct-v1.5† | 0.2347 | 0.1503 | 0.1235 | 0.0925 | 0.1154 |
| Magicoder-6.7B-S-DS† | 0.2175 | 0.1171 | 0.0856 | 0.0413 | 0.1154 |
| Llama-3-8B-Instruct† | -1.0641 | -1.6825 | -2.8893 | -4.8447 | 0.0769 |
| StarCoder2-15B-Instruct† | 0.1766 | 0.1128 | 0.0799 | 0.0405 | 0.1154 |
| Phi-3-14B-128k-Instruct† | 0.2343 | 0.1605 | 0.1194 | -0.3136 | 0.1538 |
| Codestral-22B-v0.1† | 0.725 | 0.6504 | 0.5419 | 0.3756 | 0.5385 |
| PySLICoder (SMP) SFT | 0.7567 | 0.6925 | 0.6585 | 0.4974 | 0.6154 |
| PySLICoder (SMP) SFT + ITES | 0.809 | 0.7245 | 0.6858 | 0.5561 | 0.6154 |
| PySLICoder (LING) SFT | 0.5818 | 0.4871 | 0.4443 | 0.3492 | 0.1538 |
| PySLICoder (LING) SFT + ITES | 0.5371 | 0.4601 | 0.4103 | 0.3218 | 0.1923 |
| PySLICoder SFT ITES ensemble | 0.8656 | 0.8181 | 0.7882 | 0.7123 | 0.6538 |
| PySLICoder SFT ITES + Codestral ensemble | **0.9063** | **0.8665** | **0.8330** | **0.7709** | **0.8077** |
| **GPT-4†** | **1** | **0.98** | **0.94** | **0.83** | **1** |

Table 1: Performance of various LLMs on single sound law induction on Tangkhulic data. †- indicates zero shot performance.

| Model | Reward@1 | Reward@3 | Reward@5 | Reward@10 |
|---|---|---|---|---|
| Magicoder-6.7B-S-DS† | 0.0133 | -0.0052 | -0.0282 | -0.135 |
| Codestral-22B-v0.1† | 0.4461 | 0.3333 | 0.2389 | 0.0027 |
| DeepSeekCoder-33B-Instruct† | 0.0205 | 0.002 | -0.0291 | -0.1826 |
| PySLICoder (SMP) SFT | 0.1101 | 0.029 | -0.0477 | -0.2293 |
| PySLICoder (SMP) SFT + ITES | 0.1623 | 0.0391 | -0.0394 | -0.2104 |
| PySLICoder (SMP) ITES + Codestral ensemble | 0.4461 | 0.3381 | 0.267 | 0.1525 |
| GPT-4† | 0.5568 | 0.399 | 0.2565 | -0.3 |
| **All LLMs ensemble** | **0.6124** | **0.5056** | **0.4276** | **0.2985** |

Table 2: Performance of various LLMs on multi-sound laws subsets of Huishu. †- indicates zero-shot performance.

| Model | Reward@1 | Reward@3 | Reward@5 | Reward@10 |
|---|---|---|---|---|
| Luo's MCTS** | **0.3205** | -0.3489 | -1.0175 | -0.9284 |
| Codestral-22B-v0.1† + ITES | 0.2065 | 0.1768 | 0.1603 | 0.1314 |
| PySLICoder (SMP) SFT + ITES | 0.1694 | 0.1298 | 0.1071 | 0.0732 |
| PySLICoder (SMP) SFT + Codestral ensemble ITES | 0.2104 | **0.1853** | **0.1678** | **0.1435** |
| GPT-4† + ITES* | 0.1249 | 0.1089 | 0.1041 | 0.088 |

Table 3: Performance of various LLMs on final reflex prediction. We use 20 beams with all the LLMs and show the averaged metrics across all 5 languages. †- indicates zero-shot performance. We use example selection for all LLMs. * - indicates the results are on a subset of the data. ** - We utilize 10 beams for Austronesian languages with Luo and 20 beams for Huishu.

We speculate that the poor performance of LLMs on Huishu and Tongan is respectively due to higher number of rules and more complex rules (in terms of the environment/predicates) required to explain the sound changes (through manual annotation we found 49 BasicActions to explain most of the sound changes in Huishu, while for Tongan several rules required environments with 4 or 5 predi-cates with complex phone groupings). For Luo's poor performance on Niue, we speculate that it might be due to fewer set of rules required (5 BasicActions in our annotations). Scaling to large rule cascades like those required by Huishu is a limitation of our sound law beam search algorithm and the synthetic data which is targeted towards single law prediction. The latter falls in line with the observations made by (Li and Ellis, 2024) that LLMs do well on programming by example when fine-tuned with data belonging to the same distribution as the test data.

## 6 Discussion

In this work, we present a novel programming by examples (PBE) formulation of the task of sound law induction for diachronic linguistics. Additionally, we also proposed an approach for leveraging LLMs as code synthesizers for sound law induction under this formulation motivated by their success in code generation from test cases and examples in past work (Austin et al., 2021; Lahiri et al., 2022; Jain et al., 2022). However, we found that most open-source LLMs fail to perform this PBE task even for the relatively simple, "single law prediction" where only one sound change needs to be induced from the examples, while larger more advanced models like GPT-4 can handle them with relative ease.

To tackle this issue we proposed novel language agnostic synthetic data generation algorithms for fine-tuning the LLMs on task-specific data which can greatly improve their performance in line with observations found in recent work on benchmarking LLMs for PBE (Li and Ellis, 2024). However, limitations of this approach, especially when it comes to generalization to more complicated settings of inducing multiple sound laws (multi-law synthetic reflex prediction and final reflex prediction) become apparent as evidenced by the steady drop in performance across the two tasks (Table 2 and Table 3).

When compared to traditional methods like Luo's MCTS (Luo, 2021) we find that the LLMs largely get outperformed by them especially for languages that require long rule cascades like Huishu. However LLMs can complement such approaches for some languages like Niue where smaller rule cascades can explain the sound changes. This is also points to the limitations of our sound law beam search algorithm and the importance of having search algorithms that can leverage hierarchical planning like Luo's MCTS.

## 7 Conclusions and Future Work

We propose a novel formulation of sound law induction as programming by examples (PBE) and leverage the programming ability of LLMs to propose a sample efficient solution for sound law induction in low resource settings. We observe that most open source LLMs perform poorly on this task but can be improved by fine-tuning on synthetic language agnostic PBE data, for which we provide multiple algorithms for creating data with varying levels of realism.

However we notice that advantages of this fine-tuning are limited to predicting a single sound law from examples and multi-sound law induction remains a challenge. Additionally traditional methods that can leverage hierarchical planning are more equipped to handle scenarios that require long sound law cascades to explain the sound changes. This suggests that future work should explore more hierarchical sound law search methods that can more effectively decompose sound changes and induce sound laws and combine them. Additionally it should be explored how much synthetic data for multi sound law prediction can help. At the same time the results also show that LLMs can complement traditional methods in some cases (like Proto Austronsian to Niue in our data)

### Limitations

- **Error Propagation:** A notable challenge in our method is the potential for errors to compound during multi-step sound law induction. Initial inaccuracies may escalate, leading to significant errors in the final results.

- **Computational Demands:** The use of Large Language Models (LLMs) requires considerable computational resources. Both fine-tuning and running LLMs demand extensive memory and GPU capabilities.

- **Context Length Constraints:** LLMs are limited by the amount of information they can process at once, with a maximum context length of around 4000 tokens. Prompts exceeding 50 words may be truncated, posing a limitation to our approach.

## Ethics Consideration

In developing Large Language Models (LLMs) for sound law induction, it is imperative to ensure that the synthetic data utilized for fine-tuning is devoid of biases that could distort the generated sound laws. This necessitates a careful consideration of linguistic diversity and the avoidance of overrepresentation of specific language families or phonetic patterns. Moreover, using pre-existing linguistic data or frameworks in creating synthetic data must be approached with respect for intellectual property rights, ensuring proper attribution and consent from original researchers. Lastly, the automation of sound law induction presents a transformative potential for historical linguistics, promising to alleviate cognitive and temporal demands. However, balancing the benefits of automation and the invaluable role of human expertise and interpretative analysis in the field is crucial, safeguarding the nuanced understanding of linguistic evolution.

## A  Method

### A.1  BasicAction Class Definition

The BasicAction class is used to represent the phonological transformations behind the sound laws. These transformations move from left to right over the list of tokenized phones (where the start and end are marked by a special token '#' and separated by a special token '@') in a word to match an environment specified as a list of predicates, where each predicate is a boolean function that matches to a set of phones (e.g. the is_nothing function matches to the '@' separator token). When a match is found deletions, substitutions (replacing phones with other phones based on a mapping), or insertions (mapping/replacing a phone with multiple phones) are performed at the set of relative positions specified in the list of change_pos according to the corresponding mapping function passed in the list mapping_fn.

An example BasicAction object showing the format of the predicates, change positions, and mappings is shown in Figure 2. An important implementation detail of the BasicAction is suppression of self-feeding (when rules can create the environment for their own application) which is achieved by a two-stage process of first detecting all locations/sites where an environment match is found and the modifying the appropriate phones in the second stage.

## B  Experiments

### B.1  Rules in Tangkhulic Data

Table 4 shows all the 26 rules in the Proto-Tangkhulic to Tangkhulic data used for single sound law prediction.

| Language | Sound Law | Environments | Mappings |
|---|---|---|---|
| Huishu | a→e / _ j | aj | ej |
| Huishu | a→e / _ # | a# | e# |
| Huishu | a→o / _ ŋ | aŋ | oŋ |
| Huishu | a→o / _ k | ak | ok |
| Huishu | a→o / _ w | aw | ow |
| Huishu | b→v / # _ <br> bʷ→v / # _ | #b, #bʷ | #v, #v |
| Huishu | ɨ→u | ɨ | u |
| Huishu | ʃ→s / # _ | #ʃ | #s |
| Huishu | ∅→k / i _ # | i# | ik# |
| Huishu | ∅→k / u _ # | u# | uk# |
| Huishu | k→ʔ / _ # <br> p→ʔ / _ # <br> t→ʔ / _ # | k#, p#, t# | ʔ#, ʔ#, ʔ# |
| Huishu | a→∅ / a _ | aa | a |
| Huishu | a→∅ / Consonant (C) _ | Cw | C |
| Kachai | d→ð / # _ | #d | #ð |
| Kachai | hw→f / # _ | #hw | #f |
| Kachai | ∅→w / k _ u <br> ∅→w / k _ o <br> ∅→w / k _ a | ku, ko, ka | kwu, kwo, kwa |
| Kachai | pʷ→w / # _ | #pʷ | #w |
| Kachai | e→∅ / _ j <br> o→∅ / _ j | ej, oj | j, j |
| Kachai | h→∅ / _ l <br> h→∅ / _ r | hl, hr | l, r |
| Kachai | l→∅ / _ not word init (W) | lW | W |
| Kachai | w→∅ / e _ # | ew# | e# |
| Kachai | t͡s →ð / # _ | #t͡s | #ð |
| Ukhrul | a→ɐ / not word init (W) _ not (word init and vowel) Wv | WaWv | WɐWv |
| Ukhrul | a→∅ / a _ | aa | a |
| Ukhrul | d→r | d | r |
| Ukhrul | w→∅ | w | ∅ |

Table 4: Sound laws in the single sound law induction Tangkhulic dataset.

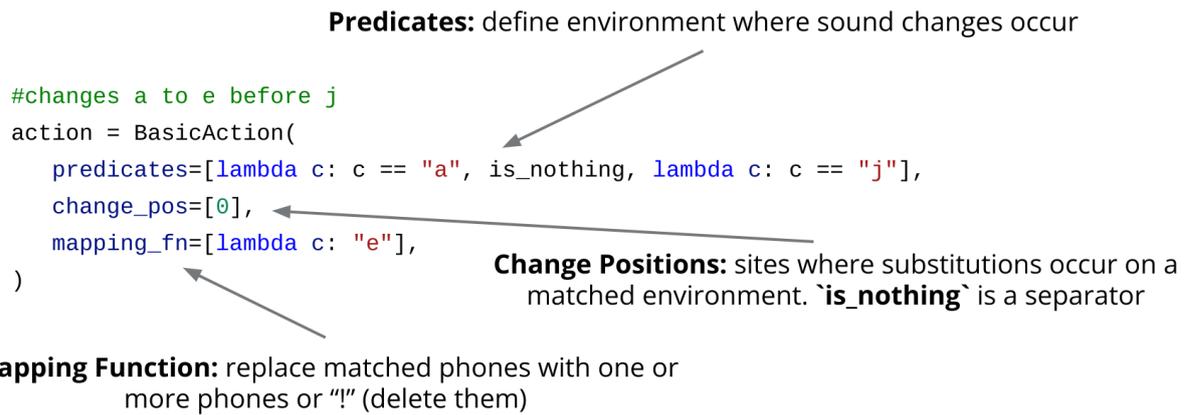

Figure 2: Instantiation of the BasicAction class to represent a sound law. This example shows a rule where "a" goes to "e" when it occurs before a "j". The predicates match to an environment of "a@j" where '@' is the separator, and then the first character of the environment or "a" goes to "e" as described by the change_pos and mapping_fn. In other words it represents the rule $a > e \backslash \_j$

### B.2 Synthetic Data Generation

Algorithm 1 shows the pseudo-code/procedure for generating the string manipulation (SMP) sound laws data while Algorithm 2 shows the procedure for generating the linguistically motivated data (LING).

---
**Algorithm 1** Synthetic Data Generation Algorithm for String Manipulation Sound Laws (SMP)
---
1: **procedure** GENERATESOUNDLAWS() ▷
2:     $e \leftarrow$ CHOICE($[1, 2, 3], w = [0.7, 0.2, 0.1]$)   ▷ sample environment size (E) based on weights (w)
3:     $BC \leftarrow$ CHOICE($[S, E, NS, NE, Null], w = [\frac{1}{16}, \frac{1}{16}, \frac{1}{16}, \frac{1}{16}, \frac{3}{4}]$) ▷ boundary condition (BC): start (S), end (E), not start (NS), not end (NE) or no condition (Null).
4:     $P \leftarrow$ SAMPLE($\mathcal{P}, k = E$)     ▷ sample k(=E) predicates (P) from set of possible predicates ($\mathcal{P}$)
5:     $nc \leftarrow$ SAMPLEINTEGER($1, E$)     ▷ sample number of changes (NC) made by sound law
6:     $CP \leftarrow$ SAMPLEINTEGERS($1, E, k = nc$)   ▷ sample nc change positions (CP) within predicates
7:     $F_M \leftarrow$ SAMPLEOPS($add, del, sub, k = nc$)   ▷ sample operations (phone addition, deletion or substitution) to create mapping function $F_M$
8:     $SL \leftarrow$ CREATEBASICACTION($P, CP, F_M$)   ▷ construct BasicAction for sound law
9:     **return** $SL$
10: **procedure** GENERATEPROTOFORMSANDREFLEXES($SL, n = 50$)   ▷ sound law (SL), number of examples (n)
11:     $Pro \leftarrow [\ ]$
12:     **for** $\frac{3}{5}N$ iterations **do**   ▷ Sample completely random words
13:         $Pro$.INSERT(RANDOMSTR())
    ▷ Sample random words having the environment as prefix or suffix
14:     **for** $\frac{1}{10}N$ iterations **do**
15:         $Pro$.INSERT($SL.P$.STR() + RANDOMSTR())   ▷ Convert predicates to string
16:     **for** $\frac{1}{10}N$ iterations **do**
17:         $Pro$.INSERT(RANDOMSTR() + $SL.P$.STR())
    ▷ Sample random words having the environment in the middle
18:     **for** $\frac{1}{10}N$ iterations **do**
19:         $Pro$.INSERT($SL.P$.STR() + RANDOMSTR() + $SL.P$.STR())
20:     **for** $\frac{1}{10}N$ iterations **do**
21:         $Pro$.INSERT($SL.P$.STR() + RANDOMSTR() + $SL.P$.STR() + RANDOMSTR() + $SL.P$.STR())
22:     $Ref \leftarrow$ APPLYSOUNDLAW($SL, Pro$)   ▷ Apply sound law on all protoforms to generate reflexes
23:     **return** $Pro, Ref$
---

**Algorithm 2** Synthetic Data Generation Algorithm for Linguistic Sound Laws (LING)

1: **procedure** GENERATENONCES()  ▷ create nonce words
2:     $langs \leftarrow [nld, fra, deu, ita, pol, spa, vie]$
3:     **for** $lang \in langs$ **do**
4:         $lex \leftarrow$ WUGGYLEXICON($lang$)  ▷ retrieve lexicon from Wuggy
5:         $lex_{1000} \leftarrow$ RANDOMSAMPLE($lex$, 1000)  ▷ reduce lexicon size to 1000 by random sampling
6:         $nonces_{lang} \leftarrow$ [WUGGYNONCEWORD($x$) | $x \in lex_{1000}$]  ▷ Use Wuggy to find a nonce word for each lexeme
7:     **return** [$nonces_{lang}$ | $lang \in langs$]
8: **procedure** CREATERULE()  ▷ create synthetic sound law (to be used in synthetic language creation)
9:     **for** $ctx \in [PreContext, SeqToChange, PostContext]$ **do**
10:         $len_{ctx} \leftarrow \lfloor |\text{SAMPLEGAUSSIAN}() + 1| \rfloor$  ▷ sample length of sequence
11:         **for** $len_{ctx}$ iterations **do**  ▷ loop through phones to make requirements
12:             **for** $feat \in$ ARTICULATORYFEATURES() **do**  ▷ requirements on level of articulatory feats.
13:                 $sample \leftarrow$ SAMPLEGAUSSIAN()  ▷ sample for requirements
14:                 **if** $sample \leq -1$ **then**  ▷ require $feat == 0$...
15:                     REQUIRE($ctx, iteration, feat == 0$)
16:                 **if** $sample \geq 1$ **then**  ▷ ...or $feat == 1$ with equal probability
17:                     REQUIRE($ctx, iteration, feat == 1$)  ▷ ...though no requirement is most likely
18:     **for** $phone \in$ RANGE($len_{SeqToChange}$) **do**
19:         **if** CHOICE($[0, 1], w = [\frac{1}{8}, \frac{7}{8}]$) $== 0$ **then**  ▷ $\frac{1}{8}$ chance
20:             DELETE($phone$)  ▷ ...of deletion
21:         **if** CHOICE($[0, 1], w = [\frac{1}{8}, \frac{7}{8}]$) $== 0$ **then**  ▷ $\frac{1}{8}$ chance
22:             $sub \leftarrow$ CREATESUBFEATURES()  ▷ For a substitution, follow the same process as when assigning requirements (feature by feature), but using CHANGETO($feat, x$) instead of REQUIRE($feat == x$)
23:             SUBSTITUTE($phone, sub$)  ▷ ...of substitution
24:         **if** CHOICE($[0, 1], w = [\frac{1}{16}, \frac{15}{16}]$) $== 0$ **then**  ▷ $\frac{1}{16}$ chance
25:             $rand \leftarrow$ RANDOMPHONE()
26:             INSERTBEFORE($phone, rand$)  ▷ ...of insertion
27:         **if** CHOICE($[0, 1], w = [\frac{1}{16}, \frac{15}{16}]$) $== 0$ **then**  ▷ $\frac{1}{16}$ chance
28:             $rand \leftarrow$ RANDOMPHONE()
29:             INSERTAFTER($phone, rand$)  ▷ one last chance to insert after
30:     **return** $rule$  ▷ rule consists of all requirements, deletions, substitutions, and insertions
31: **procedure** CREATELANGUAGES()
32:     $langs \leftarrow [\ ]$
33:     **for** $NumLangs$ iterations **do**
34:         $BaseLang \leftarrow$ CHOICE($[nld, fra, deu, ita, pol, spa, vie]$)
35:         $protos \leftarrow$ RANDOMSAMPLE($nonces_{BaseLang}$, 50)  ▷ sample 50 protoforms from nonces
36:         $laws \leftarrow [\ ]$
37:         **while** LENGTH($rules$) $< NumRules$ **do**  ▷ $NumRules = 3$ by default, i.e. 3 rules per synthetic language
38:             $rule \leftarrow$ CREATERULE()
39:             **if** APPLIES($rule, protos, 3$) **then**  ▷ if the rule applies to at least protos
40:                 $rules$.INSERT($rule$)
41:         $sudolang \leftarrow (rules, protos)$  ▷ synthetic langauge consists of rules and protoforms
42:         $langs$.INSERT($sudolang$)
43:     **return** $langs$

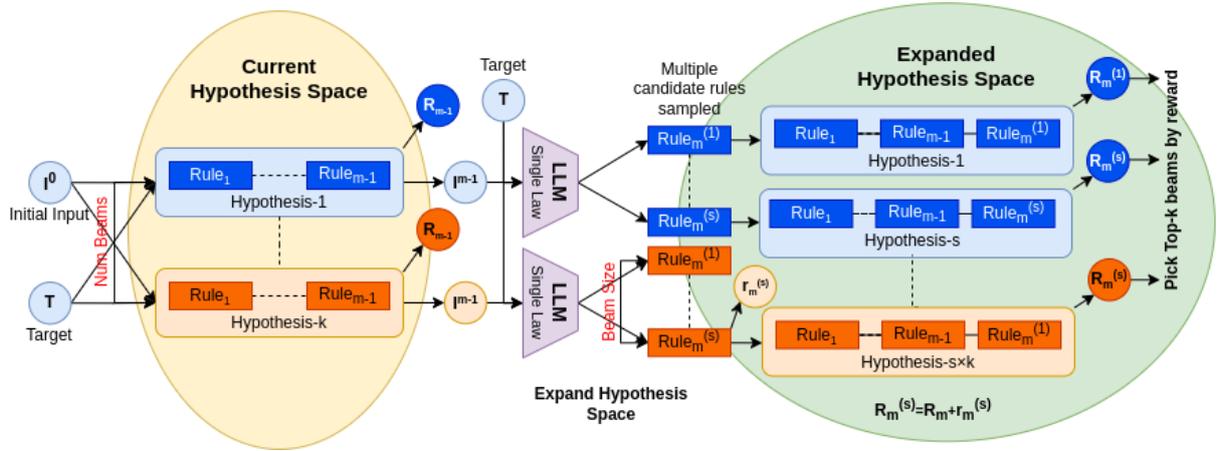

Figure 3: Sound law beam search algorithm for uncovering rule cascades with cumulative edit distance reward as the guiding heuristic

| Model | | Reward@1 | Reward@3 | Reward@5 | Reward@10 |
|---|---|---|---|---|---|
| Luo's MCTS | Niue | -0.6149 | -1.0621 | -2.2298 | -1.6876 |
| | Tongan | 0.3841 | -0.8344 | -2.0795 | -1.7298 |
| | Samoan | 0.3515 | -0.3766 | -1.0990 | -0.9498 |
| | Hawaiian | 0.6596 | 0.1518 | -0.0482 | -0.2838 |
| | Huishu | 0.822 | 0.3766 | 0.3691 | 0.0088 |
| Codestral-22B-v0.1+ ITES | Niue | 0.1553 | 0.1387 | 0.1329 | 0.118 |
| | Tongan | 0.0728 | 0.0596 | 0.053 | 0.0424 |
| | Samoan | 0.2899 | 0.2563 | 0.2345 | 0.2008 |
| | Hawaiian | 0.3347 | 0.3079 | 0.2788 | 0.2178 |
| | Huishu | 0.1796 | 0.1215 | 0.1022 | 0.0779 |
| PySLICoder (SMP) SFT + ITES | Niue | 0.1484 | 0.1227 | 0.1176 | 0.0967 |
| | Tongan | 0.0674 | 0.0674 | 0.0629 | 0.0494 |
| | Samoan | 0.163 | 0.1389 | 0.1239 | 0.0986 |
| | Hawaiian | 0.2065 | 0.1896 | 0.1623 | 0.1138 |
| | Huishu | 0.0288 | 0.0192 | 0.0159 | 0.0119 |
| GPT-4 + ITES | Niue | 0.1374 | 0.1374 | 0.1374 | 0.1176 |
| | Tongan | 0.1124 | 0.0805 | 0.0708 | 0.0584 |

Table 5: Breakdown of final reflex prediction task results per descendant/observed language. We ran inference with 5, 10, 20, 25, and 50 max steps of the beam search respectively for Niue, Tongan, Samoan, Hawaiian, and Huishu. We restricted the GPT-4 experiments to Niue and Tongan due to the high API cost of running inference for the larger Samoan, Hawaiian, and Huishu datasets.